\documentclass[letterpaper, 10 pt, journal, twoside]{IEEEtran}
\usepackage{graphics} % for pdf, bitmapped graphics files
\usepackage{epsfig} % for postscript graphics files
\usepackage{times} % assumes new font selection scheme installed
\usepackage{amsmath} % assumes amsmath package installed
\usepackage{amssymb}  % assumes amsmath package installed
\usepackage{url}
\ifCLASSINFOpdf
\else
\fi
\begin{document}
\title{Learning Structured Representations of \\ Spatial and Interactive Dynamics for \\ Trajectory Prediction in Crowded Scenes}

% \author{Michael~Shell,~\IEEEmembership{Member,~IEEE,}
%         John~Doe,~\IEEEmembership{Fellow,~OSA,}
%         and~Jane~Doe,~\IEEEmembership{Life~Fellow,~IEEE}% <-this % stops a space
% \thanks{M. Shell was with the Department
% of Electrical and Computer Engineering, Georgia Institute of Technology, Atlanta,
% GA, 30332 USA e-mail: (see http://www.michaelshell.org/contact.html).}% <-this % stops a space
% \thanks{J. Doe and J. Doe are with Anonymous University.}% <-this % stops a space
% \thanks{Manuscript received April 19, 2005; revised August 26, 2015.}}
\author{Todor Davchev$^{\dagger*}$ Michael Burke$^{\ddagger}$ and Subramanian Ramamoorthy$^{\dagger}$%
\thanks{Manuscript received: May, 15, 2020; Revised August, 13, 2020; Accepted December, 06, 2020.}%Use only for final RAL version
\thanks{This paper was recommended for publication by Tamim Asfour upon evaluation of the Associate Editor and Reviewers' comments.
Davchev is supported by an EPSRC Industrial CASE award with Thales Maritime Systems. Ramamoorthy would like to acknowledge support from EPSRC ORCA Hub (EP/R026173/1).)} %Use only for final RAL version
\thanks{$^{*}$ Corresponding Author.}%
\thanks{$^{\dagger}$ Davchev and Ramamoorthy are with the School of Informatics, University of Edinburgh, 10 Crichton St, EH8 9AB, United Kingdom, 
        {\tt\small \{t.b.davchev, s.ramamoorthy\}@ed.ac.uk}}%
\thanks{$^{\ddagger}$ Burke is with ECSE, Monash University, 14 Alliance Lane, Clayton Campus, Melbourne, Australia, 3800, {\tt\small michael.burke1@monash.edu}, Work done when Burke was with University of Edinburgh}
\thanks{Digital Object Identifier (DOI): see top of this page.}
}

\markboth{IEEE Robotics and Automation Letters. Preprint Version. Accepted December, 2020}
{Davchev \MakeLowercase{\textit{et al.}}: Learning Representations of Spatial and Interactive Dynamics for Trajectory Prediction} 




\maketitle

\begin{abstract}
Context plays a significant role in the generation of motion for dynamic agents in interactive environments. This work proposes a modular method that utilises a learned model of the environment for motion prediction. This modularity explicitly allows for unsupervised adaptation of trajectory prediction models to unseen environments and new tasks by relying on unlabelled image data only. We model both the spatial and dynamic aspects of a given environment alongside the per agent motions. This results in more informed motion prediction and allows for performance comparable to the state-of-the-art. We highlight the model's prediction capability using a benchmark pedestrian prediction problem and a robot manipulation task and show that we can transfer the predictor across these tasks in a completely unsupervised way. The proposed approach allows for robust and label efficient forward modelling, and relaxes the need for full model re-training in new environments.
\end{abstract}

% \begin{IEEEkeywords}
% IEEE, IEEEtran, journal, \LaTeX, paper, template.
% \end{IEEEkeywords}

% \fi
\IEEEpeerreviewmaketitle

\section{Introduction}
% for IEEE journal papers produced under \LaTeX\ using
% IEEEtran.cls version 1.8b and later.
\IEEEPARstart{A}{utomated} decision making requires long-term predictions of the behaviours of surrounding agents. For example, an autonomous vehicle requires knowledge of the future positions of any surrounding pedestrians if it is to plan successful paths. Similarly, a robotic arm may need to know the future position of its surrounding agents (i.e. another arm) for collaborative planning. This paper investigates predictive models that address this challenge.

A commonly adopted approach is to rely on labelled data that is often expensive to obtain and thus existing solutions tend to rely on a few well known datasets for training \cite{gupta2018social,pfeiffer2018data}. Generalising to completely new tasks using models trained on these such datasets, however, is challenging due to the complexity of group dynamics and environment specific motion. Real-world motion is dependent on varying environmental cues and informal rules of interaction. For example, groups of people often tend to walk in similar directions, on pavements, or through choke points like bridges. Nonetheless, the underlying local motion remains very similar across many tasks. 

This work proposes a modular model-based approach that allows each of these aspects to be learned from data in a decoupled fashion. This decoupling allows the re-use of individual modules, thus reducing labelling and training requirements. We take the view that it is better to adapt a model to environment specific dynamics by conditioning on scenes than to train a global model across all possible dynamics. For example, rather than re-training a large model to perform motion prediction in a new scene by relying on labelled data, we propose to re-learn only parts of the model while transferring existing dynamics components.  
        
More formally, we consider models trained on trajectory-based motion prediction in supervised settings and the unsupervised transfer of these prediction models. We consider the case where environments are observed through a fixed camera feed and have a history of labelled, agent positions in associated frames. In the absence of any knowledge of the long term behaviour of an observed agent, even making simple local predictions about the general direction of motion is non trivial. This work focuses on learning effective state representations utilised in a forward model that can successfully infer a distribution over the long-term behaviour of a given agent.
        
We introduce a forward model comprising three key components. First, the model makes use of an environment-specific spatial encoding (\textbf{R}), that captures information about the existing objects in a scene. This information is then used within a global dynamics component (\textbf{D}), which models the evolution of this scene over time. We condition \textbf{D} on the 2D locations of all agents present in the scene. This provides us with a hidden state, or a summary of the scene dynamics that takes into account the moving components therein. We use the spatial encoding along with the hidden summary to train a local (per given agent) behavioural model (\textbf{B}). The proposed architecture (RDB) explicitly incorporates stochastic elements in each module, so as to allow a measure of uncertainty in the predictions. This uncertainty can be used in downstream tasks requiring prediction. The modularity of RDB allows for component re-use across tasks.

The global dynamics model (D) learns to model the evolution of the scene in latent image space (R), capturing information about where motion typically occurs. This can be learned in either a weakly supervised or fully unsupervised manner as it requires no ground truth labelling. The local dynamics model (B) is trained (using position labels) to predict future positions conditioned on this information. Importantly, this also allows for domain transfer, as we can condition an existing dynamics model to adapt to new environments and tasks, by simply retraining the global (D) and environment specific (R) models.

We show through an ablation study that the three modules (R, D, B) are complimentary and collectively contribute to achieving the highest performance, outperforming the respective sub-components. This shows that decoupling environmental from the local per-agent dynamics requires an additional, global dynamics component in order to preserve the effective multi-agent state prediction.
\begin{figure*}%[thpb]
\vspace{1mm}
      \centering
      \framebox{\includegraphics[width=0.96\textwidth]{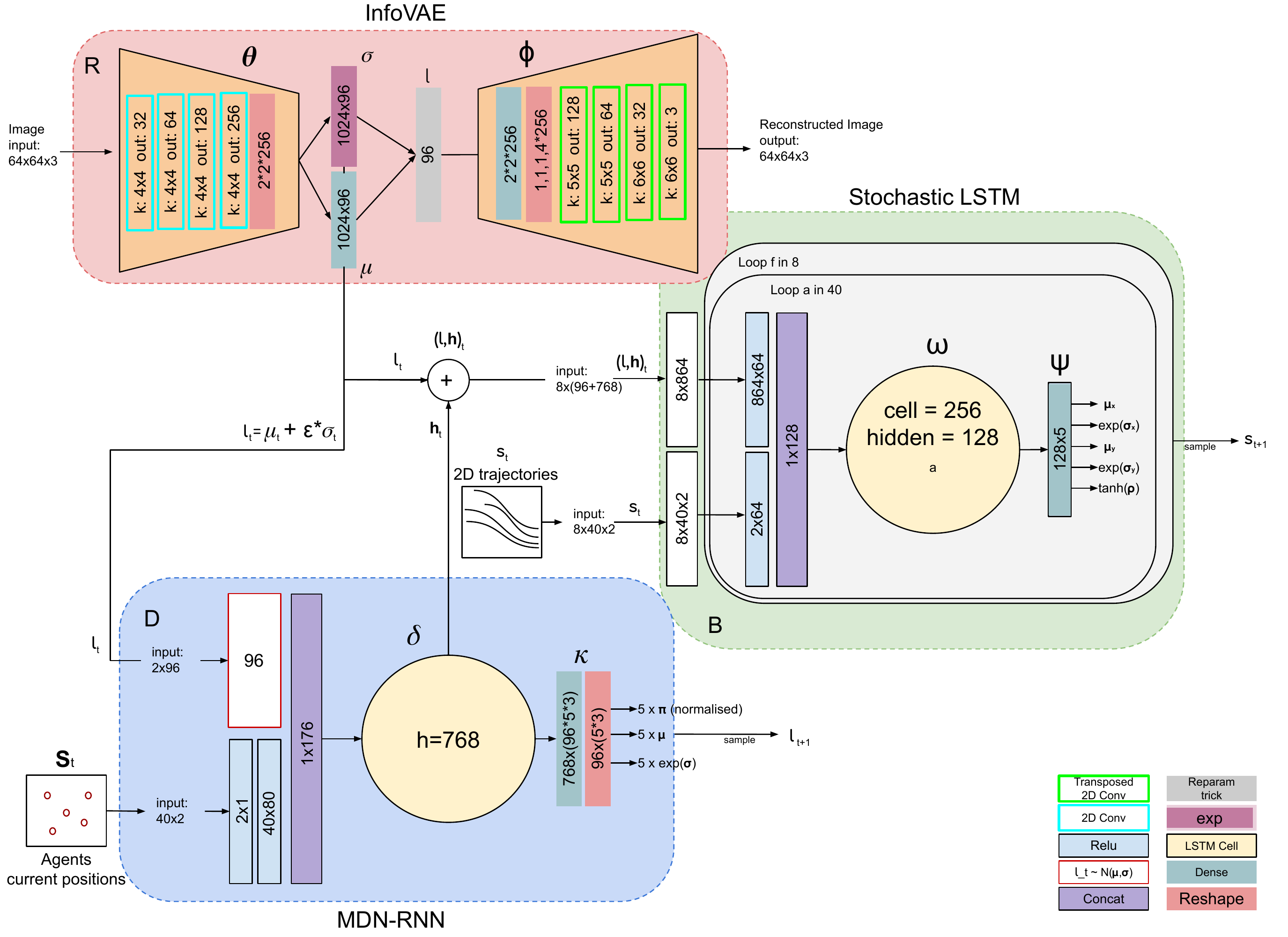}
}
      \caption{An illustration of the proposed model. Each of the three components R, D, B are trained separately in the listed order. Here, \textbf{R} is the spatial representation module which takes in as input an image $I_t$ and obtains a latent representation $l_t$, trained using an InfoVAE loss \cite{zhao2017infovae}. \textbf{D} is the module responsible for modelling the evolution of the entire scene, i.e. the global dynamics. It takes in as input the latent representation from \textbf{R}, $l_t$, along with a vector $S_t$ comprised of the Cartesian positions of all agents at time $t$ and learns to predict a distribution over potential spatial latent representations $l_{t+1}$ for time $t+1$. \textbf{B} models individual agents' ($a$) motion. It takes in as input $s_t^a$ along with $l_t$ from \textbf{R} and the obtained summary $h_t$ from \textbf{D}. }
      \label{fig:architecture}
    %   \vspace{-7mm}
\end{figure*}
We highlight the flexibility of the proposed model through experiments conducted on crowded scene tasks and a tabletop manipulation scenario, and by illustrating unsupervised local dynamics transfer between these entirely different domains and tasks relying on unlabelled image data only. Here, we show that the proposed model is able to learn a global representation of the environment and use this to predict motion following a learned plan in a never seen before configuration. In summary, the main contributions of this paper\footnote{\url{https://sites.google.com/view/rdb-agents/}} are:
\begin{itemize}
    \item a modular architecture that allows for a variety of effective domain adaptation schemes for model-based motion prediction tasks;
    % \item to show that decoupling the extraction of an environment specific representation from motion prediction allows for better adaptation to new tasks.
    \item showing that both spatial and global dynamic aspects of environments are key to building effective representations for trajectory prediction; and
    \item a novel modular and label efficient architecture for model-based agent trajectory prediction in the context of both crowded and structured scenes.
\end{itemize}

\section{Related Work}
\label{sec:related_work}
\textbf{Model-based prediction}: There has been significant interest in sequence generation and model-based learning. Model-based learning is used widely across robotics applications, particularly when it comes to collaborative tasks and modelling state-space representations to enable control of an individual agent~\cite{8931564, ha2018recurrent}. Recently, there has been increasing interest in model-based prediction methods for crowded scene prediction tasks~\cite{rudenko2020human} .
Pfeifer et al.~\cite{pfeiffer2018data}, Hasan et al.~\cite{hasan2018mx} and Lisotto et al.~\cite{lisotto2019social} make use of semantic spatial and occupancy representations for trajectory prediction. This contrasts with our approach which uses an image representation trained in an unsupervised fashion. A combination of local and global, environmental features was previously explored in \cite{xue2018ss} and global multiagent dynamics are used in~\cite{zhao2019multi}. However, those methods do not model scenes as stochastic processes and rely on labels to model the interactions between agents. Ballan et al.~\cite{ballan2016knowledge} introduce a Dynamic Bayesian Network which exploits the scene specific knowledge for trajectory prediction in a stochastic manner and shows that such knowledge can be transferred across similar tasks. We build upon these works by studying ways to obtain environmental models in an unsupervised way. By separating the stochastic environmental models from the learned predictors, we allow for transfer of those learned predictors across a range of different tasks, for example, moving  a model trained for crowded scene prediction to a structured robot action prediction task.

Recent work extends the notion of probabilistic trajectory prediction by utilising a variational approach to learn static scene representations in an end-to-end fashion \cite{li2019conditional}, demonstrating the benefits of employing a generative world model. Liang et al. \cite{liang2019peeking} build on this by proposing to jointly learn two complimentary tasks in an end-to-end system using rich visual features for human trajectory and activity prediction. We build upon these works by decoupling the generative spatial representation from the model by learning an additional, complimentary global dynamics component. We show that this achieves performance comparable to state-of-the-art while enabling transfer of the learned task-specific predictors across different tasks.

\textbf{Sequence Generation}: Sequence modelling using neural distributions has been widely adopted in a number of domains. Graves et al.~\cite{graves2013generating} uses recurrent neural networks for text and handwriting analysis. Here, plausible future samples are drawn from a Gaussian distribution and a mixture density network (MDN) \cite{bishop1995neural} is used to generate real-valued data. Ha et al.~\cite{ha2017neural} use a similar model to generate sketches, relying on a temperature parameter $\tau$ to control model uncertainty during inference. Alahi et al.~\cite{alahi2016social} propose an approach that builds upon \cite{graves2013generating} and models social interactions through a pooling layer that uses the latent state representations of surrounding agents. 

In the context of multi-agent pedestrian modelling, social dynamics are often used to facilitate prediction. For example, the Social Forces (SF) method, \cite{helbing1995social}, uses a potential field based approach to model the social interactions between various pedestrians. Lerner et al.~\cite{lerner2007crowds} use an example-based reactive approach to model pedestrian behaviour by creating a database of local spatio-temporal examples. Pellegrini et al.~\cite{pellegrini2009you} model the behaviour of pedestrians in crowded settings using linear extrapolation over short intervals via a Linear Trajectory Avoidance (LTA) technique. Complimentary solutions that successfully capture pedestrian behaviours in specific situations \cite{kuderer2012feature,kitani2012activity} have also been proposed, but these methods rely on additional semantic information. Lee et al.~\cite{lee2017desire} enable trajectory generation by scoring a number of possible paths by a learned criterion. This concept was also explored using Generative Adversarial Networks (GANs)~\cite{gupta2018social} and further extended through the work of Sadeghian et al.~\cite{sadeghian2019sophie} who combine GANs with physical constraints via learned spatial attentive mechanisms. Radwan et al.~\cite{radwan2018multimodal} use a dilated CNN, jointly optimising the trajectories of all agents in a scene to leverage the inter-dependencies in motion without the need for explicit modelling. In contrast, we focus on extracting environmental features in an unsupervised and weakly supervised fashion as part of a modular architecture. We show that in doing so we achieve performance comparable to state-of-the-art and allow for unsupervised transfer across different tasks.

Graph Convolutional Networks (GCN) have recently gained popularity in crowded scene trajectory prediction. Ivanovic et al.~\cite{ivanovic2019trajectron} use dynamic spatiotemporal graphs, Kosaraju et al.~\cite{kosaraju2019social} capture the multi-modal nature of human trajectories by combining a graph convolutional network with a GAN in an end-to-end fashion, while Mohamed et al. \cite{mohamed2020social} represent the social interactions between pedestrians and their temporal dynamics using a spatio-temporal graph. Although GCNs show significant promise, they have yet to be explored for transfer. 

\textbf{Summary}: While model-based pedestrian prediction has been studied extensively in the past, to date there have been no investigations into unsupervised model transfer across domains. The primary contribution of this paper is a modular architecture that allows for unsupervised adaptation to new domains. By structuring our model into 3 components, an environment embedding (R) learned in an unsupervised fashion, a local dynamics module (B) learned using supervision (labels available), and a global dynamics model (D) that integrates the two and which can be trained in either an unsupervised or supervised fashion, we allow for the transfer of models trained in different settings to both similar and entirely new domains. 

\section{Methodology}
\label{sec:methods}
This work's focus is on predicting the future motion of an arbitrary number of observed agents (i.e. their behaviour) whose action spaces and objectives are unknown. More specifically, we aim to predict the two-dimensional motion of agents in video sequences. We assume we are given a history of two-dimensional position annotations and video frames as a sequence of RGB images. 
Each agent $a$ ($a \in [1 \ldots N]$, where $N$ is the maximum number of agents in a given sequence) is represented by a state ($s_t^a$) which comprises xy-coordinates at time $t$, $s_t^a = (x, y)_t^a$. The global world state ($S_t$) is then defined as a tuple comprising the set of all agents' locations in that frame ($S_t = \{(\bigcup_{a=1}^{N} s_t^a)\}$), for the scene at time $t$ represented by an image $I_t$. Given a sequence of $obs$ observed states $S = \{S_{t-obs}, S_{t-obs+1}, \ldots, S_{t-1}, S_{t} \}$, we formulate the prediction as an optimisation process, where the objective is to learn a posterior distribution $P(Y | S, I)$, over multiple agents' future trajectories $Y$. Here an individual agent's future trajectory is defined as ($\hat{s}^a = \{s_{t+1}^a, s_{t+2}^a, \ldots, s_{t+pred}^a\}$) for $pred$ steps ahead for every agent $a$ found in scene $I_t$.

Even though we do not have access to agents' objectives and available actions spaces, we hypothesise that the decision making process for each agent is influenced by the behaviour of the surrounding agents. Additionally, there may be spatial objects that could act as regions of interest across the given environment, affecting an agent's motion. Moreover, it is often the case that all agents follow some informal, common rules of motion or dynamics that apply globally to the entire scene. We model these spatial and dynamic components separately and from the perspective of an observer.

To this end, we build probabilistic generative models of real world environments and use these to help predict the locations of those agents at multiple instances in the future. We refer to this technique as model-based prediction. In particular, we model agent behaviours using three components; \textbf{R}- an autoencoder that embeds the scene image $I_t$ into low dimensional representations $l_t$, \textbf{D}- a global dynamics component that models the transition between two scenes in the same low dimensional representation space $l_t$ to $l_{t+1}$ by utilising the latent vectors extracted from \textbf{R}; and \textbf{B} - a conventional stochastic LSTM that takes in as input the 2D positions of the agents as well as the representations obtained by \textbf{R} and \textbf{D}. We discuss each of these components in more detail below.

The first, a visual sensory component (\textbf{R}), encodes the environment into a latent vector $l_{t}$. This vector then acts as the global state representation. This latent state representation can be useful if it can preserve information about the existing agents at time $t$ in terms of position with respect to the surrounding environment and additional directional queues. In this work, we assume that such information is preserved if the model can successfully reconstruct the surrounding environment as well as the existing agents at a given timestep.
However, conventional variational approaches can fail to reconstruct individual agents due to their changing locations and small size in a video (see additional experiments on website$^{1}$). To overcome this, we use an information maximisation variational autoencoder (InfoVAE) which matches the moments of encoded and decoded images \cite{zhao2017infovae}.
   
The representation module \textbf{R} is trained by minimising the distance between the input image and its decoded variant using a maximum mean discrepancy (MMD) \cite{gretton2007kernel} loss:
\begin{multline*}
      R: \min_{\theta, \phi} \mathcal{L}(I_t, \theta,\phi) = \text{MMD}(q_\phi(l_t)||p(l_t))
      + \\ \mathbb{E}_{p_{data}(I_t)}\mathbb{E}_{q_{\phi}(l_t|I_t)}[p_{\theta}(I_t|l_t)]
\end{multline*}

Here, $\theta, \phi$ are the neural model parameters, $I_t$ is the RGB image scene at time step $t$ where each image $I_t \in \mathbf{I}_t \subseteq R^{n \times j}$ is an frame of a video sequence $\mathbf{I}$. $l_t$ denotes the latent image representation and $p(\cdot) \sim N(0, \mathbf{I})$ is a standard Normal distribution. $p_{\theta}$ is a generative distribution parametrised as a neural network, and  $q_{\phi}$ a distribution mapping images into the latent space, also represented as a neural network (see Figure~\ref{fig:architecture}). $p_{data}$ denotes the unknown target distribution from which images in the dataset are sampled from.

To learn the second module \textbf{D}, we train an RNN-MDN conditioned on agent positions using the image embedding obtained from \textbf{R}, $l_t$. This input is conditioned on a fixed length 2D vector comprised of the locations of all existing agents $S_t$ at time $t$. The network learns to model the distribution over the next possible latent vector $l_{t+1}$ in the latent space learned by \textbf{R} and provides us with a summary $h_{t}$ that encodes the temporal dynamics of the global environment (See Figure~\ref{fig:architecture}). We sample predictions for $l_{t+1}$ from a multivariate factored mixture of Gaussians parametrised by a learned mean $\mu$ and variance $\sigma$ and a mixing coefficient $\pi$. We assume a diagonal covariance matrix of a factored Gaussian distribution to simplify computation. We define summary $h_0 \rightarrow 0,$ and then every other $h_t = f_{\delta}(h_{t-1}, (l_t, S_t)),\ \forall t \in (0, T]$, for a dynamics function $f(\cdot)$ represented with an LSTM. We then obtain the summary $h_t$ from $f(\cdot)$ that maximises the likelihood $g_{\kappa}(h_t) = p(l_{t+1} | l_t, S_t)$, where $g(\cdot)$ is a feedforward neural network with weights and biases denoted by $\delta, \kappa$ and $l_t$ is a latent representation of image $I_t$ situated in the latent space obtained by \textbf{R}. This results in the loss function: 
%\vspace{-1.5mm}
\begin{multline*}
    D: \min_{\delta, \kappa} \mathcal{L}(l_{t+1}, \delta, \kappa) \\=  \sum_{t=1}^{T} - log \big(\pi_{t}\mathcal{N}(l_{t+1}|\mu_{\leq t}, \sigma_{\leq t}; \delta, \kappa) \big)
\end{multline*}
When sampling at runtime, we adjust a temperature parameter $\tau$ over the mixture weights to control model uncertainty \cite{jang2016categorical} (See the supplementary material on the website$^1$ for demonstrations of this). Adjusting $\tau$ is useful when training the \textbf{B} component of the proposed model. We use $\tau$ to to remedy over-fitting, and allow for a probabilistic model. Adjusting $\tau$ controls the amount of randomness in \textbf{D} and thus allows for manual uncertainty calibration~\cite{ha2017neural}. Higher values for $\tau$ result in larger variability while lower leads to more deterministic predictions. Temperature close to 0.0 tends to  predict only the static cues of the environment while temperature close to 1.0 is more likely to predict agents appearing at random locations, while values around the middle tend produce more reasonable predictions.

Finally, we utilise both \textbf{R} and \textbf{D}, along with each acting agent's history of motion to predict future positions using temporal prediction module (\textbf{B}). This fuses the local dynamics of individual agents with those of the global scene (represented with a tuple ($l_t$, $h_t$), see Figure~\ref{fig:architecture}). We iterate over each agent separately  which allows the network to be independent of the number of agents which proves useful for reusing \textbf{B} across tasks. We define \textbf{B} as an LSTM $f_{\omega}(\cdot)$, parametrised by $\omega$, followed by a feedforward layer, parametrised by $\psi$, that predicts the parameters of a single 2D Gaussian distribution over agent position, with mean $
    \mu$, variance $\sigma$ and correlation $\rho$. B then minimises:
\begin{multline*}
    B: \min_{\omega, \psi} \mathcal{L}(Y, \omega, \psi) = \\ \sum_{t=1}^{T} - log \bigg(\sum_{j}\mathcal{N}(s_{t+1}^j|\mu^{j}_{t\leq}, \sigma_{t\leq}^{i}, \rho_{t\leq}^{i}; \omega, \psi )\bigg)
\end{multline*}
    where $s_t^i = (x, y)_t^i$ is the $i^{th}$ agent's predicted Cartesian position, sampled from the learned 2D Gaussian distribution. We found the inclusion of the cross-correlation term $\rho$ between the $x$ and $y$ positions of the future steps of an agent important for the tasks considered here.
    
    We learn each of these components separately, assigning most of the network capacity to \textbf{R} and \textbf{D} while restricting \textbf{B} to a small network trainable on CPU. Further, we optimise with respect to an expectation of the individual losses over the distribution individual samples were drawn from.

\section{Experimental Evaluation}
The proposed architecture is evaluated using multiple publicly available datasets for predicting human behaviour in crowded scenes as well as on a task introduced in this work, which reflects the applicability of the proposed solution to a robotic setting. Each experiment is conducted by training \textbf{D} and \textbf{B} on $n-1$ environments, while the evaluation is performed on another, previously unseen environment. In this test environment, we re-train only the unsupervised spatial encoding module \textbf{R} on a subset of the test sequence. All results are then generated using the remaining data in the test environment, which had not been seen by any of the network modules. 
\begin{table*}
\label{tabl:motion_prediction}
\begin{center}
\caption{Reported results assume observation lengths of $T_{obs}=1.6sec$ and prediction lengths of $T_{pred}=3.2sec$ and $T_{pred}=4.8sec$. We report the Average and Final Displacement Errors in normalised pixel space. Overall, using environmental models works better while our modular architecture performs on-par with the considered alternatives whilst doing better on FDA.}
\begin{tabular}{|l||r|r|r|r|r|r|r|}  
\hline
$T_{pred}=3.2s$  &  Model \textbackslash Dataset & ETH HOTEL & ETH UNIV & UCY UNIV & UCY ZARA1 & UCY ZARA2 & AVERAGE\\
\hline
\hline
 & B-LSTM & 0.069 / 0.148 & 0.067 / 0.130 & 0.056 / 0.114 & 0.079 / 0.177 & 0.050 / 0.110 & 0.064 / 0.136 \\
\hline
Average / Final & S-LSTM \cite{alahi2016social} & 0.077 / 0.150 & 0.111 / 0.219 & 0.078 / 0.159 & 0.072 / 0.152 & 0.060 / 0.119 & 0.080 / 0.160 \\
\hline
Displacement Error & SNS-LSTM \cite{lisotto2019social} & \textbf{0.033} / 0.132 & \textbf{0.030} / 0.127 & 0.065 / 0.253 & \textbf{0.023} / 0.100 & \textbf{0.023} / 0.090 & \textbf{0.035} / 0.140 \\
\hline
 & RDB & 0.041 / \textbf{0.072} & 0.054 / \textbf{0.110} & \textbf{0.054} / \textbf{0.108} & 0.038 / \textbf{0.073} & 0.038 / \textbf{0.076} & 0.046 / \textbf{0.088}\\
\hline
\hline
$T_{pred}=4.8s$ & Model \textbackslash Dataset & ETH HOTEL & ETH UNIV & UCY UNIV & UCY ZARA1 & UCY ZARA2 & AVERAGE \\
\hline
\hline
 & B-LSTM & 0.116 / 0.248 & 0.098 / 0.172 & 0.092 / 0.196 & 0.144 / 0.338 & 0.092 / 0.216 & 0.108 / 0.234 \\
\hline
Average / Final & S-LSTM \cite{alahi2016social} & 0.132 / 0.255 & 0.169 / 0.296 & 0.129 / 0.279 & 0.122 / 0.269 & 0.098 / 0.206 & 0.130 / 0.261\\
\hline
Displacement Error & SNS-LSTM \cite{lisotto2019social} & \textbf{0.035} / 0.198 & \textbf{0.030} / 0.182 & \textbf{0.081} / 0.453 & 0.029 / 0.169 & \textbf{0.025} / 0.136 & \textbf{0.040} / 0.228 \\
\hline
 & RDB (ours) & 0.057 / \textbf{0.103} & 0.083 / \textbf{0.143} & 0.088 / \textbf{0.188} & 0.060 / \textbf{0.123} & 0.061 / \textbf{0.129} & 0.070 / \textbf{0.137} \\
\hline
 \hline
\end{tabular}
\end{center}
\vspace{-5mm}
\end{table*}
\subsection{Training Schedule}
All hyper-parameters were optimised using grid search applied over independent video segments sampled at random from training datasets (see web page for more details). Frames were re-sized to $64\times64$ pixels, and were further processed using contrast limited adaptive histogram equalisation (CLAHE) \cite{zuiderveld1994contrast} which enhances the contrast in tiles that are populated with agents. 
While training B, we consider all agents in a sequence as a single mini batch. This reduces the variance of the parameter updates with respect to the particular scene at hand, and also leads to more stable convergence$^{1}$.

\subsection{Motion Prediction on Crowd Datasets}
The first experiment conducted uses the ETH Hotel, ETH University \cite{pellegrini2009you} and UCY University, Zara1 and Zara2 \cite{lerner2007crowds} datasets. %Each of these is recorded at 25 frames per second (fps) while the data acquisition is applied at 2.5fps. 
The datasets comprise $\approx 4000$ frames and contain $\approx 1600$ agents that follow both linear and non-linear trajectories. They include agents walking on their own, in social groups or passing by one another. %We assume that we know in advance the locations of all agents at each time step and that we have access to the video arrays in an RGB format. 
We combine these datasets, following previous work \cite{alahi2016social}, \cite{radwan2018multimodal}, and apply a leave-one-out procedure during training for both dynamic components \textbf{D} and \textbf{B}. 

\begin{table}
\centering
\caption{Ablation table on the use of spatial ($R$) and scene dynamics ($D$) for predicting agent trajectories with $B$ (see Sec. \ref{sec:methods}).}
\label{tabl:ablation}
\resizebox{0.48\textwidth}{!}{
\begin{tabular}{|l||r|r|r|r|}
\hline
% Dataset \textbackslash Model &   B   & R+B   &  D+B  & R+D+B \\
Dataset \textbackslash Model & B$(s_t)$   & B$(s_t, l_t)$  & B$(s_t, h_t)$  & B$(s_t, l_t, h_t)$ \\
\hline
ETH HOTEL                    & 0.069 / 0.148 & 0.049 / 0.089 & \textbf{0.041} / \textbf{0.072} &\textbf{ 0.041} / \textbf{0.072} \\
\hline
ETH UNIV                     & 0.067 / 0.130 & 0.069 / 0.129 & 0.060 / \textbf{0.110} & \textbf{0.054} / \textbf{0.110} \\
\hline
UCY UNIV                     & 0.056 / 0.114 & 0.065 / 0.131 & 0.061 / 0.123 & \textbf{0.054} / \textbf{0.108} \\
\hline
UCY ZARA1                    & 0.079 / 0.177 & 0.065 / 0.130 & 0.044 / 0.084 & \textbf{0.038} / \textbf{0.073} \\
\hline
UCY ZARA2                    & 0.050 / 0.111 & 0.053 / 0.106 & 0.043 / 0.083 & \textbf{0.038} / \textbf{0.076} \\
\hline
AVERAGE                      & 0.064 / 0.136 & 0.060 / 0.117 & 0.050 / 0.094 & \textbf{0.046} / \textbf{0.088} \\
\hline
\end{tabular}}
\vspace{-5mm}
\end{table}
We benchmark the performance of the proposed approach against several methods for motion prediction and namely; the social S-LSTM \cite{alahi2016social}, SNS-LSTM \cite{lisotto2019social} and a vanilla LSTM with stochastic outputs (B-LSTM).

In order to report fairly, we used the recommended training settings for these models and report results in Table I. %\ref{tabl:motion_prediction}.

% Comparison methods are trained and evaluated using the proposed data pre-processing and test pipeline, but employ provided model implementations. 
We evaluate the accuracy of the motion prediction models using the following metrics:
\begin{itemize}
    \item Average Displacement Error: that is the root mean squared error (RMSE) over all predicted positions from a single pedestrian's trajectory.
    \item Final Displacement Error: that is the RMSE of the last predicted location from a single pedestrian's trajectory.
\end{itemize}
Each network is trained using a length of 8 frames and during inference using observation length of 4 (1.6s) and prediction lengths of 8 frames (3.2s) and 12 frames (4.8s). Observing a smaller portion of the trajectory makes it more challenging for a model to predict future behaviour, especially in a scenario where the network has been trained with sequences smaller than 4.8 seconds as in our case.

\begin{figure}
\vspace{1mm}
\centering
\includegraphics[width=0.486\textwidth]{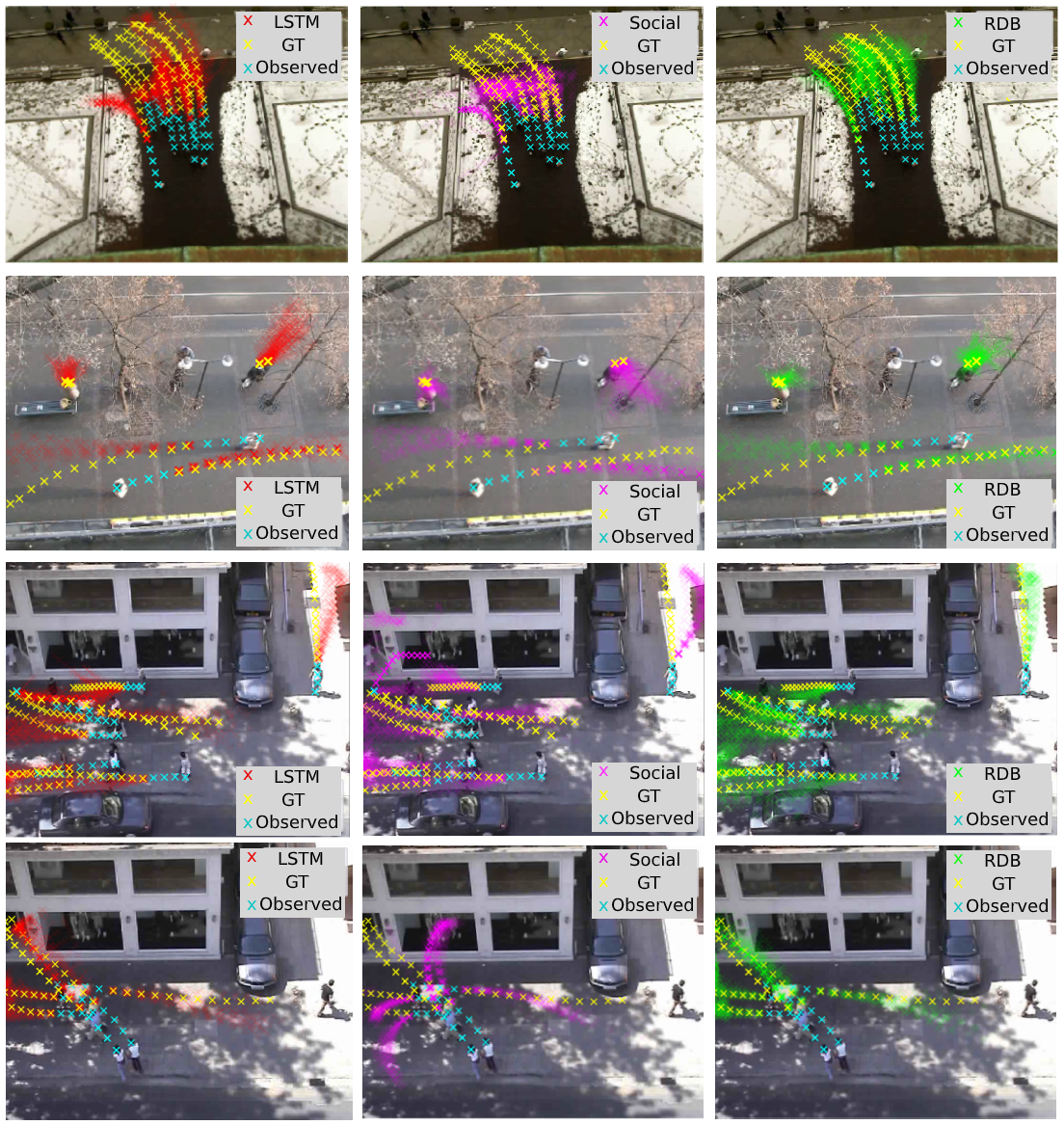}
\caption{A comparison of model performance in crowded scenes. The first two rows show results on ETH UNIV and HOTEL \protect\cite{pellegrini2009you}, while the remainder are from UCY ZARA1 and ZARA2 \protect\cite{lerner2007crowds}. An agent's trajectory is tracked for 4 steps (light blue) and is then predicted for the next 12 (ground truth depicted in yellow). Rows represent different dataset and columns - different models. Red depicts a simple LSTM with a stochastic output, magenta - modelling only social interactions between agents \protect\cite{alahi2016social} and green are the results from this work. RDB takes  environmental cues into consideration and maintains more consistent predictions.\label{fig:crowded_scenes}}
\vspace{-7mm}
\end{figure}

Table I shows the results obtained on this task. In all considered cases, utilising environmental properties (as both this work and SNS-LSTM~\cite{lisotto2019social} do) of a data set outperforms alternative approaches. This is particularly the case when we consider end position predictions that are highly dependent on both the static representations and the natural flow of motion in a given dataset. For example, for motion going up the road as in UCY Zara1 (see Figure~\ref{fig:crowded_scenes}), we see a significant difference between RDB and the remaining models.
Lisotto et al. ~\cite{lisotto2019social} incorporate environment information using semantic maps generated from the reference image. While this is effective, it can limit transfer to different domains where semantic classification models are not available. In contrast RDB uses a latent image representation that is learned in an entirely unsupervised fashion to incorporate contextual environment information into the trajectory prediction. This results in performance comparable to state-of-the-art in terms of displacement error and improved predictions in terms of FDE.

\subsubsection{Ablation Study \& Qualitative Analysis}
An ablation study was performed in order to evaluate the importance of each of the components of the proposed model. %Note that the B component of our network is exactly the same as a vanilla LSTM. 
Here we consider the impact of trajectory prediction using each of the learned representations in Figure~\ref{fig:architecture}.

Specifically we consider the cases where dynamics model \textbf{B} only has access to positions ($s_t$); when it is provided with both positions and the instantaneous scene representation ($s_t,l_t$) learned using \textbf{R}; and finally, when it receives positions, instantaneous scene representations learned using \textbf{R}, and the dynamic scene representation ($s_t,l_t,h_t$) learned using \textbf{D}.
Table \ref{tabl:ablation} shows that using a combination of both global dynamics and static representations is most effective.

A key finding of this table is that the global scene dynamics module \textbf{D} is particularly important, and that just adding an image representation to a monolithic dynamics model \textbf{B} $(s_t, l_t)$ is insufficient. Instead, only by conditioning the local dynamics model on the scene dynamics, do we gain the benefits of this architecture. 

Figure~\ref{fig:crowded_scenes} further highlights these observations. In the first row, groups of people cross a bridge. Groups are often moving together and continue walking on the pavement. Further, none of the people walks over the bridge or steps on the snow. Such cues are omitted in the models that do not have input representations that take this into account. Further, predicting no motion with such techniques is often difficult. Row two shows a scenario in which 4 out of the 6 people stay in the same place at a train station. While all three models considered do relatively well at predicting the behaviour of the two moving people, only RDB manages to predict that the rest of the agents stand still at a train stop. The model also utilises environmental cues in the remaining datasets, where it successfully anticipates that agents are walking on the road or are walking towards the door of the store.

These results not only reiterate the point that a good latent representation of the static environment is required for prediction, but also demonstrate the flexibility of a modular solution. In addition to retaining near state-of-the-art performance, a key benefit of this modular architecture is that it allows for unsupervised domain transfer since it decouples the environment-specific context from the actual per-agent prediction model. We study the efficacy of this approach to transfer in the following subsection.

\subsection{Robot Action Prediction}

\begin{figure}[!h]
      \vspace{-2mm}
      \centering
      \includegraphics[width=0.486\textwidth]{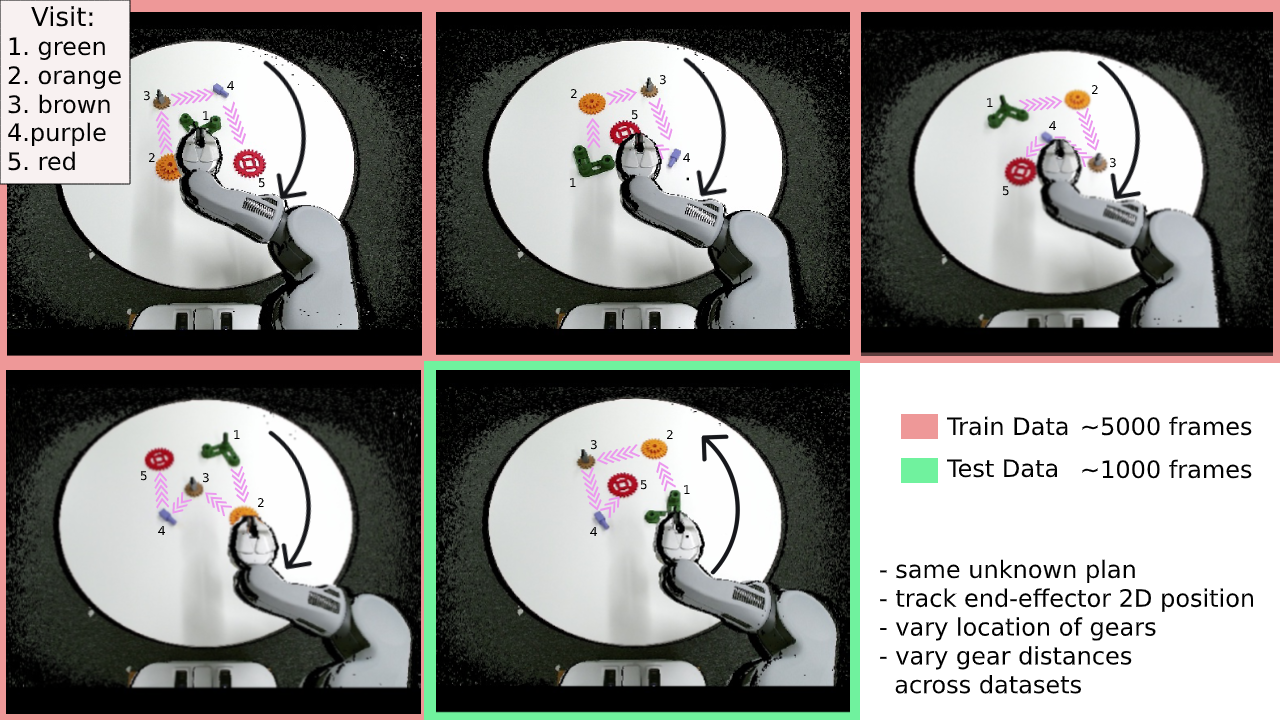}
    %   \framebox{\parbox{6.4in}{\includegraphics[width=6.4in,clip,keepaspectratio]{robo_dataset2.png}
      \caption{An illustration of the behaviours considered in the robot gripper tracking dataset. A red background indicates settings considered as the training data while a green background indicates the task used at test time. Note that the direction of motion is completely opposite at test time. See webpage$^{1}$ for larger version.}
      \label{fig:robot_task}
\end{figure}

In order to study the use of RDB for domain adaptation, we introduce a tabletop manipulation dataset and a sequence of domain adaptation tasks. %We build 5 datasets of similar length to ETH and UCY. 
Here, our goal is to predict the position of a robot gripper that follows a predetermined plan to inspect components required for an assembly task (see Figure~\ref{fig:robot_task}). 

We vary the position of each gear across different datasets, but maintain the same order of visitation, which is determined by the colour of a gear. %The total length of each dataset is approximately 1000 frames. We record the location of the robot's gripper and project this location into the image plane. 
Across all collected datasets, the required parts of the assembly are ordered in arbitrary manner. However, their inspection order remains the same. %In all examples, we begin by visiting the green gear and end at the red one as described in Figure~\ref{fig:robot_task}. 
We consider orderings that allow for a clockwise inspection of the gears  during training and monitor performance on a never seen before item ordering, that requires anti-clockwise inspection at test time. This task includes a non-linear alteration of the data that is particularly challenging to model. It should be noted that adaptation in this task can not be achieved through strategies like data augmentation, as component position governs visitation order. We train the proposed model using four out of the five video sequences (See Figure~\ref{fig:robot_task})  using eight observation frames during training, and 12 observation frames with prediction length of 32 frames during inference. Unlike the crowded scenes, the focus of the task is to measure the capacity of the proposed approach to relate visual cues to its prediction and adapt to unseen configurations. In such settings, standard prediction models would be expected to overfit to the training data and fail to predict the intended motion correctly unless conditioned on environmental cues.
 
 \begin{figure}
 \vspace{2mm}
      \centering
      \includegraphics[width=0.4865\textwidth]{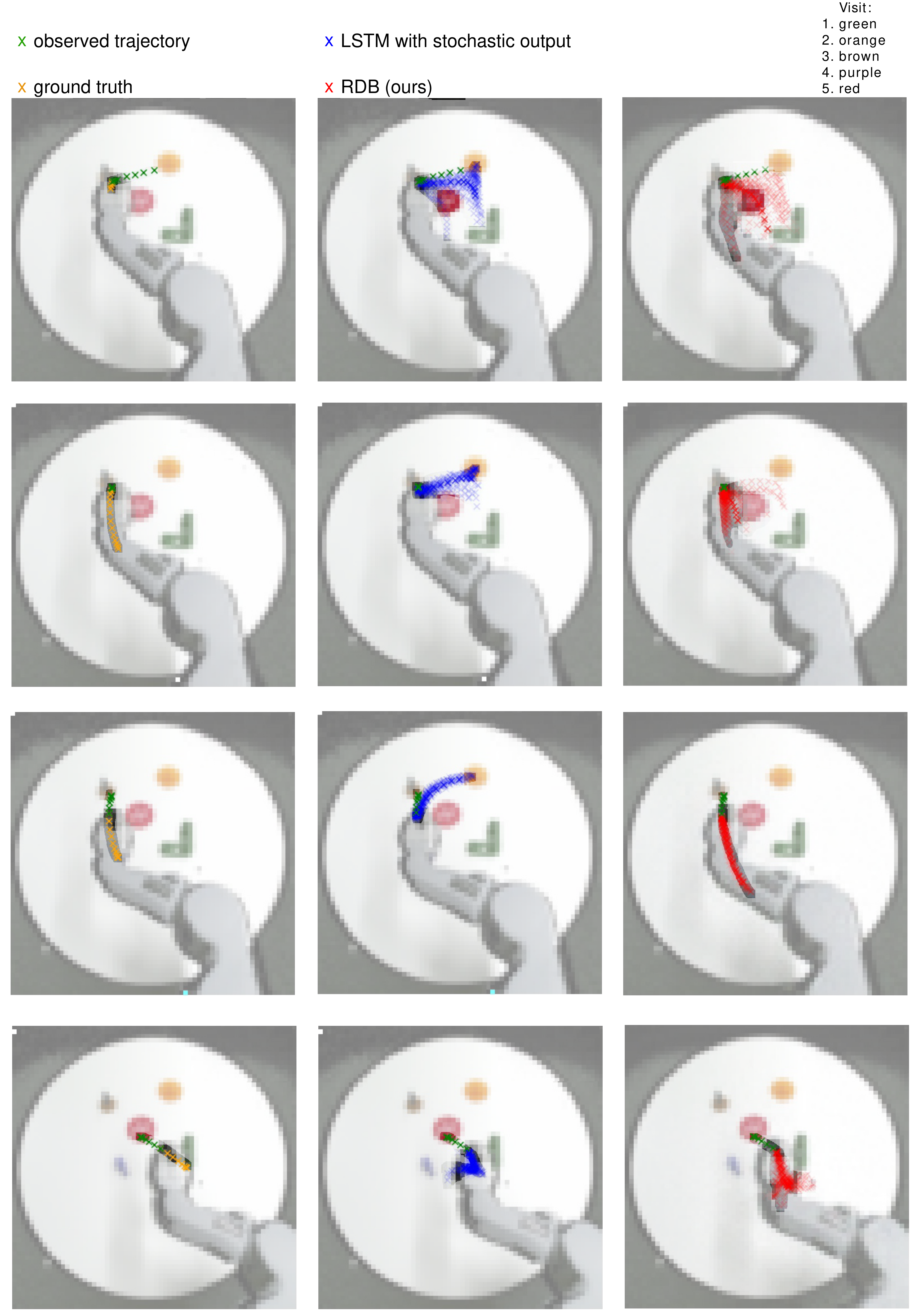}
      \caption{Examples of predicted trajectories. Green lines depict the observed trajectory, the rest are predictions. The left-most column shows ground truths, followed by predictions made with a standard LSTM and predictions using RDB. The results suggest that the proposed solution learns a weak dependency between the colour coded objects in the environment and the plan followed by the agent.} %The upper right corner indicates the unknown to the observer plan.}
      % 8x12 -> rdb: 0.08 ... just: 0.09
      \label{fig:RDB_in_action}
      \vspace{-7mm}
\end{figure}

Figure~\ref{fig:RDB_in_action} compares the performance of a standard stochastic LSTM and RDB on this task. At inference, each model is fed a sequence of the executed trajectory (depicted in green), and then tasked with predicting the future trajectory -- the ground truth depicted in yellow. Quantitative results are reported in rows 3 and 4 of Table~\ref{tabl:transfer}, in the 'Robot' column. As expected a standard LSTM overfits to the training data and predicts that the observed agent would go back towards the start of the observed trajectory (see the middle column in Figure~\ref{fig:RDB_in_action}). In contrast, RDB successfully retrieves paths that correspond to the plan followed (right-most column) resulting in three times better performance in terms of ADE. Throughout this experiment the robotic arm covers different gears during visitation. This makes it much harder to predict motion towards certain gears, since some gears will not be present in the scene at certain times. Nevertheless, the RDB model does seem to be aware of the environmental cues to a sufficient extent that it can make good use of them at prediction time. 

\begin{table}
\centering
\caption{Transfer B from a source task ($s$) to a target task ($t$).}
\label{tabl:transfer}
\resizebox{0.48\textwidth}{!}{
\begin{tabular}{|l r r |r||r|r|r|}
\hline
Agent Predictor & Spatial Model & Global Dynamics & Type \textbackslash Task & Robot & Crowd \\
\hline
\hline
% LSTM ($B^{untrained}$) & 1.49 & 2.11 \\
% \hline
% $R + D^{untrained} + B^{untrained}$ & 1.38 & 1.81 \\
Random & & & n\textbackslash a & 1.44 & 1.96 \\
\hline
\hline
$B_{targ}(s_t)$ & & & supervised & 0.28 & 0.06\\
\hline
$B_{targ}(s_t, l_t, h_t)$ & $R_{targ}(\cdot)$ & $D_{targ}(\cdot)$  \textbf{(our)} & supervised & \textbf{0.10} & \textbf{ 0.04}\\
\hline
\hline
$B_{src}(s_t)$ & & & unsupervised & 0.52 & 0.48 \\
\hline
$B_{src}(s_t, l_t, h_t)$ & $R_{src}(\cdot)$ & $D_{src}(\cdot)$ \textbf{(our)} & unsupervised & 0.85 & \textbf{0.16} \\
\hline
$B_{src}(s_t, l_t, h_t)$ & $R_{targ}(\cdot)$ & $D^{untrained}(\cdot)$ \textbf{(our)} & unsupervised & 0.49 & 0.20 \\ % 0.18
\hline
$B_{src}(s_t, l_t, h_t)$ & $R_{targ}(\cdot)$ & $D^{unsup}_{targ}(\cdot)$ \textbf{(our)} & unsupervised & \textbf{0.27} & 0.18 \\
\hline
\hline
$B_{src}(s_t, l_t, h_t)$ & $R_{targ}(\cdot)$ & $D_{targ}(\cdot)$ \textbf{(our)} & weakly supervised & \textbf{0.22} & 0.18\\
\hline
\end{tabular}}
\vspace{-5mm}
\end{table}

The first row of figures illustrates a scenario where the manipulator covers the purple gear and has been hovering over the brown component for a few steps. As a result, the RDB model is less certain and predicts three potential directions of motion, moving towards the location of a purple gear, through the middle to the red gear or going back to the orange one. %The final option is the only one which goes in a direction the network has previously seen in the training data. 
Over time, the network becomes more certain of the direction of motion after it sees a few steps in that direction. This, however, is not the case for the vanilla LSTM which has heavily overfit to the training data. 

Finally, the last row's visit to the green gear indicates the end of the underlying plan. The RDB network is uncertain what visitation might follow next, but predicts two potential directions that are situated in the direction of existing gears.

\subsection{Transferring Local Dynamics Across Tasks}
%Our evaluation showed comparable and at times improved to state-of-the-art performance; and demonstrated the importance of the global scene dynamics to decoupling environmental models from the per-agent predictors. 
A key benefit of the proposed architecture is that it allows for unsupervised domain adaptation via a number of approaches. 
%This section fuses those findings to evaluate the ability of the modular architecture to transferring those local predictors (B) across both tasks. 
In this section we evaluate domain adaptation of a model trained on a source task $src$ (either crowd or robot) and then transferred to an associated target task $targ$ (either robot or crowd). %We reuse the models used in the previous experiments and investigate a few different ways of extracting unsupervised and weakly supervised environmental models. 

The rows labelled unsupervised in Table~\ref{tabl:transfer} show these results, while columns indicate target tasks. LSTM ($B_{src}$) represents a dynamics module trained on positions using the source dataset, while $R_{src}$ and $R_{targ}$ denote instantaneous scene representations trained (unsupervised) on source and target image datasets respectively. The global dynamics module is typically trained using position information alongside image representations obtained using the R module, but can also be trained in an unsupervised fashion, by conditioning on noise instead of positions, $D^{unsup}$. We compare this with an untrained dynamics module $D^{untrained}$ as an ablation.

Importantly, results above show that by conditioning a dynamics model trained on a source domain, using representation and global dynamics modules trained in an unsupervised fashion on images taken from a target domain ($R_{targ}+D^{unsup}_{targ}+B_{src}$), we achieve trajectory performance close to that of supervised learning using position labels (LSTM $B_{targ}$) when transferring dynamics learned for pedestrians to the robot end effector tracking task. In contrast, a fully unsupervised transfer from robot to crowd outperforms its alternatives, suggesting that training on the structured straight line predictions present in the robot task leads to more accurate predictions for the crowds task, even though $D_{src}(\cdot)$ cannot model local interactions. By definition (Section~\ref{sec:methods}), module \textbf{B} is independent of the number of agents for a given task. This allows us to use \textbf{B} from the robot task to the crowds prediction one. However, \textbf{D} was dependent on the maximum accepted agents size.

Although we obtain greater performance when RDB is used in a supervised learning setting where labelled data from a target domain is available, the proposed modular approach allows for reasonably successful domain adaptation, across dramatically different tasks in the absence of labels.

\section{Conclusion}
Autonomous systems operating in multi-agent environments need effective behavioural prediction models. This paper introduces a modular architecture for multi-agent trajectory prediction given image sequences and position information. The proposed architecture, RDB, is compared to existing art on established benchmark datasets and in a domain adaptation context. This work shows that both spatial and dynamic aspects of the environment are key to building effective representations in the context of multi-agent motion prediction. As a further benefit, decoupling the learning of environment specific model allows for unsupervised transfer and domain adaptation to new environments and tasks.

Empirical evaluations along with a detailed ablation study highlight the importance of the proposed representations. The modular structure uses a model of the static environment and the global dynamics of a scene, alongside local dynamics models. Results show that these models are complimentary and necessary for successful motion prediction.

% or


% \appendices

% \section{}

\section*{Acknowledgment}

The authors would like to thank A. Srivastava, J. Viereck and M. Asenov for the valuable comments on earlier drafts.

\ifCLASSOPTIONcaptionsoff
  \newpage
\fi


\bibliographystyle{IEEEtran} % use IEEEtran.bst style

\bibliography{ral}

\end{document}